\documentclass[10pt,twocolumn,letterpaper]{article}

\usepackage{cvpr}              
\usepackage{multirow}

\usepackage[dvipsnames]{xcolor}

\definecolor{cvprblue}{rgb}{0.21,0.49,0.74}
\usepackage[pagebackref,breaklinks,colorlinks,citecolor=cvprblue]{hyperref}

\title{Style-Consistent 3D Indoor Scene Synthesis with Decoupled Objects}

\author{
Yunfan Zhang$^{1}$ \quad
Hong Huang$^{2}$  \quad
Zhiwei Xiong$^{1}$ \quad
Zhiqi Shen$^{1}$ \quad
Guosheng Lin$^{1}$ \\
Hao Wang$^{2}$ \quad
Nicholas Vun$^{1}$ \\
$^{1}$Nanyang Technological University \quad
$^{2}$The Hong Kong University of Science and Technology(Guang Zhou)
}

\begin{document}
\begin{figure}
\twocolumn[{
\renewcommand\twocolumn[1][]{#1}
\maketitle
\centering
\includegraphics[width=\linewidth]{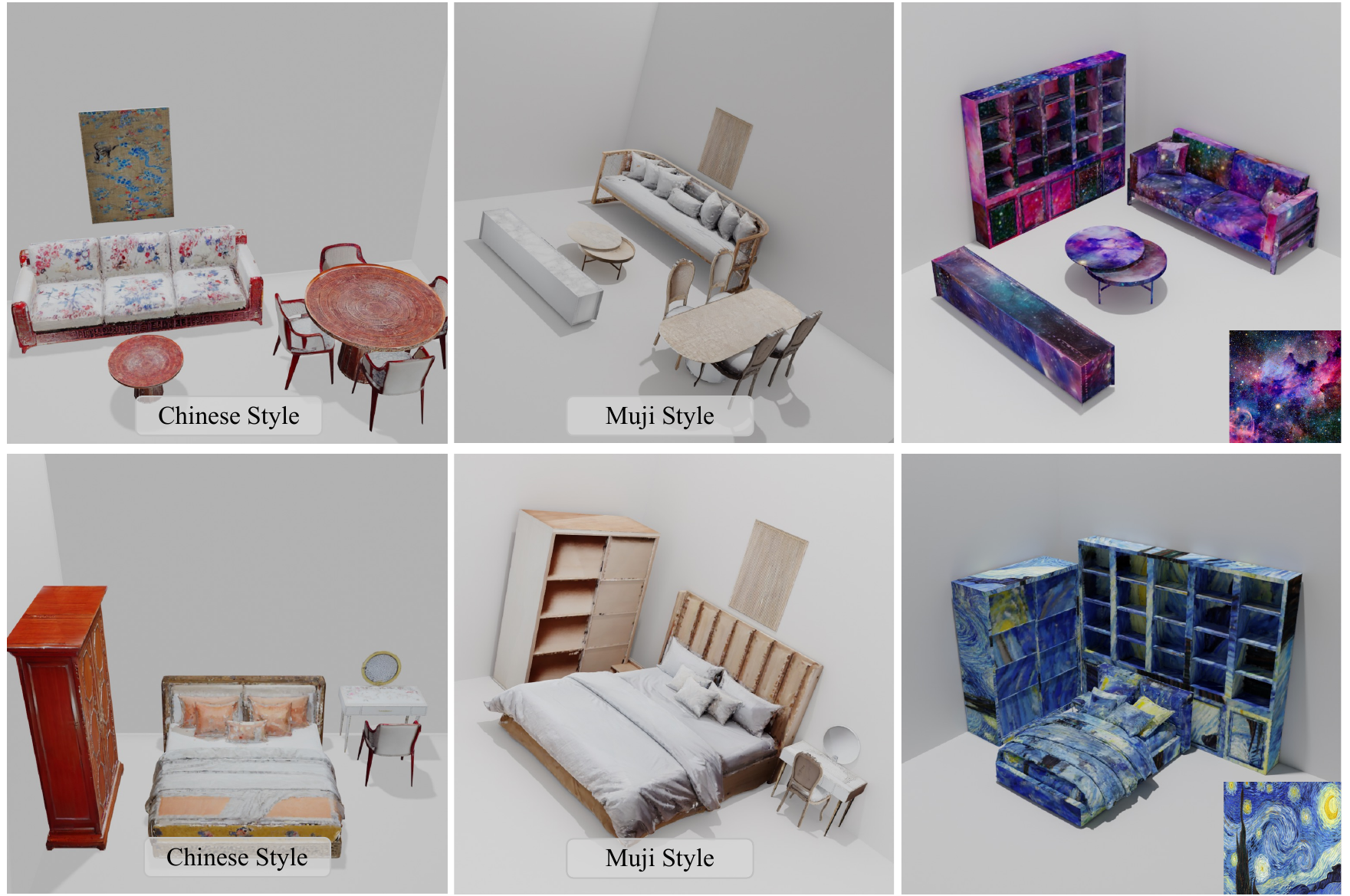}
\caption{The synthesized stylized 3D indoor scenes. The first column depicts the living room and bedroom using the text prompt \texttt{Chinese Style}; The second column depicts the living room and bedroom using the text prompt \texttt{Muji Style}; The last column depicts the living room of \texttt{Galaxy} style and bedroom of \texttt{the Starry Night} Style, in which the image prompts are used.}
\label{fig:teaser}
}]
\end{figure}

\begin{abstract}
Controllable 3D indoor scene synthesis stands at the forefront of technological progress, offering various applications like gaming, film, and augmented/virtual reality. The capability to stylize and de-couple objects within these scenarios is a crucial factor, providing 
an advanced level of control throughout the editing process. This control extends not just to manipulating geometric attributes like translation and scaling but also includes managing appearances, such as stylization. Current methods for scene stylization are limited to applying styles to the entire scene, without the ability to separate and customize individual objects.  Addressing the intricacies of this challenge, we introduce a unique pipeline designed for synthesis 3D indoor scenes. Our approach involves strategically placing objects within the scene, utilizing information from professionally designed bounding boxes. Significantly, our pipeline prioritizes maintaining style consistency across multiple objects within the scene, ensuring a cohesive and visually appealing result aligned with the desired aesthetic. The core strength of our pipeline lies in its ability to generate 3D scenes that are not only visually impressive but also exhibit features like photorealism, multi-view consistency, and diversity. These scenes are crafted in response to various natural language prompts, demonstrating the versatility and adaptability of our model.
\end{abstract}

\section{Introduction}
\label{sec:intro}
The increasing focus on high-quality indoor 3D scenes is gaining attention in academic and industrial field. This trend is particularly beneficial for advancing applications such as filming and AR/VR technologies, offering valuable insights and inspiration for both designers and consumers. Therefore, there is a critical need for an efficient approach to automatically generate high-quality 3d indoor scenes \cite{chen2022text2light, hwang2023text2scene, song2023roomdreamer, fang_ctrl-room_2023, tang_mvdiffusion_2023, hollein_text2room_nodate, yang_dreamspace_nodate}.


Indoor scenes could be represented by a 360 panorama image. Several text-driven 3D indoor scene generation approaches on a panoramic image have be explored. MVDiffusion \cite{tang_mvdiffusion_2023} incrementally generated consistent multi-view images from text prompts given pixel-to-pixel correspondences and reconstructing the 3D mesh of the room from these sub-frames, effectively addressing the typical problem of error accumulation was achieved by concurrently generating all images with a global awareness. 
Ctrl-Room \cite{fang_ctrl-room_2023} separated the modeling of layouts and appearance produce a vivid panoramic image of the room guided by the 3D scene layout and text prompt generated convincing 3D rooms with designer-style layouts and high-fidelity textures from just a text prompt.
Text2Room \cite{hollein_text2room_nodate} leverage pre-trained 2D text-to-image models to synthesize a sequence of images from different poses and then monocular depth estimation with a text-conditioned inpainting model to generate complete 3D scenes with multiple objects and explicit 3D geometry.

Implicit functions like NeRF \cite{mildenhall_nerf_2020} and tri-plane \cite{Chan2021} in 3D scene generation have also been actively explored. CC3D \cite{bahmani_cc3d_2023} represents a 2D layout-conditioned 3D generation framework, while DiscoScene \cite{xu_discoscene_2022} conditions scene generation on 3D bounding box priors.
Text2Room \cite{hollein_text2room_nodate} leverage pre-trained 2D text-to-image models to synthesize a sequence of images from different poses and then monocular depth estimation with a text-conditioned inpainting model to generate complete 3D scenes with multiple objects and explicit 3D geometry.

What's more, some works also model the whole scene using a single mesh. DreamSpace \cite{yang_dreamspace_nodate} proposed a coarse-to-fine panoramic texture generation strategy with dual texture alignment to recovery fine-grain details and authentic spatial coherence. However, these works either suffer from generating correct room layouts or fail to control the individual room objects.

To address these limitations, we propose an novel 3D indoor scene synthesis pipeline that provides multi-modal controllability, such as text prompt or images to control generation and stylization objects. This pipeline aims to synthesis 3D indoor scenes with multi-object style consistency. The key insight involves separating diverse room objects from the scene. We adopt meshes as the 3D representation, as they can be seamlessly integrated into downstream applications like AR/VR devices. They can be sourced from CAD models or generated through well-trained text-to-mesh or image-to-mesh models. Building on the capabilities of SyncDreamer \cite{liu_syncdreamer_2023}, individual mesh can be reconstructed from a single-view image, expanding the range of selectable objects significantly.


Compared to state-of-the-art 3D style transfer methods, our experiments show an improvement in terms of 3D consistent stylization both qualitatively and quantitatively. Additionally, our mesh objects representation de-couples inter-objects and object to background, allowing more degrees of freedom to manipulate explicitly.

To summarize, our contributions are:
\begin{itemize}
    \item We introduce a novel 3D indoor scene synthesis pipeline dedicated to generate de-coupled mesh objects using either text prompt or single-view images.
    \item Objects within the scenes can be stylized using either text instructions or a style image, ensuring a consistent style across multiple objects.
    \item The resulting complete indoor scenes exhibit visual coherence in both style and spatial arrangement, presenting a unified and aesthetically pleasing composition.
\end{itemize}

\section{Related Works}
\label{sec:related works}

\subsection{3D Scene Generation}
Recently, several works proposed to use different controls such as 3D bounding box, layout abstract or text prompt to generate 3D scenes. DiscoScene \cite{xu_discoscene_2022} proposed to leverage the pre-extracted 3D bounding boxes to model all objects in a scene using a single NeRF \cite{mildenhall_nerf_2020} using bounding box centre and scale as additional condition. 
CC3D \cite{bahmani_cc3d_2023} adopted the layout abstract generated from the top-down view and different color codes as the object labels to synthesis different types of objects. However, the Style-GAN \cite{karras2020training} based framework cannot fully disentangle style-code and input layouts as the layout change can result in the objects appearance change. Ctrl-Room\cite{fang_ctrl-room_2023} adopted a two-stage method to generate 3D room from text input, in which the geometric layout and appearance generation were separated. Since layout semantic panorama were generated through equirectangular projection, the generated 3D room still contains incomplete structures in invisible areas. Text2room \cite{hollein_text2room_nodate} incrementally synthesizes nearby images using a 2D diffusion model and then reconstructs depth maps to assemble these images into a 3D room model. However, it faces challenges in maintaining geometric and textural consistency among multi-posed images. MVDiffusion \cite{tang_mvdiffusion_2023} concurrently handles perspective images through a pre-trained text-to-image diffusion model. The integration of inventive correspondence-aware attention layers enhances cross-view interactions, ensuring the creation of coherent multi-view images from text prompts. This is accomplished by establishing pixel-to-pixel correspondences with a global awareness perspective.
DreamSpace \cite{yang_dreamspace_nodate} presents a novel coarse-to-fine panoramic texture generation approach for texturing the entire scene with intricate details and authentic spatial coherence. The fundamental idea is to initially conceptualize a stylized 360◦ panoramic texture from the central viewpoint of the scene and then propagate it to other areas using a combination of inpainting and imitating techniques. The model performs texture inpainting in confidential regions and subsequently utilizes an implicit imitating network to synthesize textures in occluded and small structural areas.

Note that most of these methods are restricted to well-aligned objects and fail on more complex, multi-object scene imagery. Our work instead naturally handles multi-object scenes with spatial de-coupled object-level representation. Comparison of DisCoScene and relevant works, the ability to model multiple objects in a scene and handle complex datasets beyond diagnostic scenes.

\subsection{3D Object Mesh Generation} 
Crafting high-fidelity meshes demands the skills of a seasoned professional, necessitating expertise and significant time investment. Alternatively, relying on recently pre-trained mesh generation models can yield a diverse range of generated objects, streamlining the process.

SDFusion \cite{cheng2023sdfusion} utilizes an encoder-decoder architecture to compress 3D shapes into a condensed latent representation, upon which a diffusion model is trained. However, this approach may encounter challenges in generating open-world objects due to its dependence on a limited 3D dataset. Consequently, this limitation could significantly affect its ability to handle a broader range of diverse object generation scenarios.

SyncDreamer \cite{liu_syncdreamer_2023} introduces a synchronized multi-view diffusion model that captures the joint probability distribution of multi-view images using a 3D-aware feature attention mechanism. This model is designed to maintain consistency in both geometry and colors for the generated images. 

\subsection{Neural Style Control for 3D Scene}
Text2tex \cite{chen_text2tex_nodate} utilizes a partitioned view representation by dynamically segmenting the rendered view into a generation mask. This guides the depth-aware inpainting model in generating and updating partial textures for the corresponding regions. This approach enables Text2tex to generate high-quality textures for 3D meshes based on given text prompts.

StyleMesh \cite{hollein_stylemesh_2022} improved the reconstructed mesh of a scene by optimizing a unique texture, implementing stylization across all input images simultaneously. Employing depth- and angle-aware optimization, surface normal and depth information from the mesh were used to attain a unified and consistent stylized look across the entire scene.

TEXTure \cite{yu_texture_nodate} also utilized an iterative methodology, dynamically defining a trimap to partition the rendered image into three progression states. The approach introduces a sophisticated diffusion sampling process, involving the dynamic painting of a 3D model from various viewpoints. This enables the generation of seamless textures from different perspectives. Notably, the method is versatile, as it not only generates new textures but also facilitates the editing and refinement of existing textures through the input of a text prompt or user-provided scribbles.

\begin{figure*}[ht]
\centering
\includegraphics[width=\linewidth]{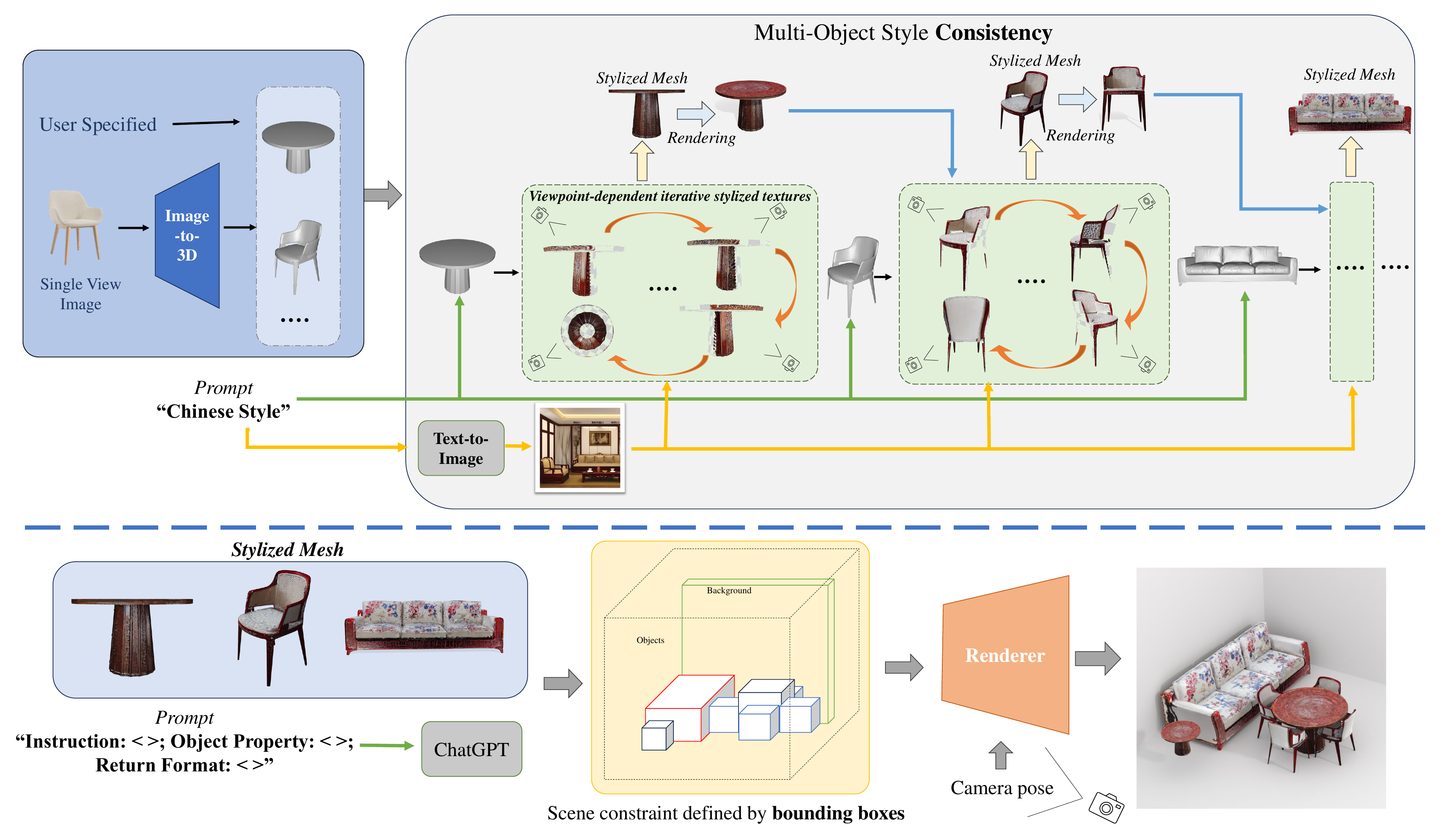}
\caption{\textbf{Model Pipeline}: Our pipeline starts by sampling objects either user specified or reconstructed from a single-view image provided by the user. Secondly, the text prompt containing style information is used to generate a styled reference scene image as global guidance. The prompt is also used to control the viewpoint-dependent stylized texturization iteratively. What's more, the previous textured mesh is used to supervise the following mesh texturization. The whole texturization is in a cascaded manner to achieve the multi-object style consistency. Subsequently, the objects are positioned and scaled within the scene based on ChatGPT learnt positions reasoning. Finally, the final scene is composed. The result could be visualized by rendering the resultant mesh using the specified camera pose.}
\label{fig:pipeline}
\end{figure*}

\section{Methods}
Our aim is to synthesis high-fidelity 3D indoor scenes featuring distinct objects with consistent styles. However, applying style transfer to the already synthesized panoramic texture treats the scene as a unified entity, making it difficult to individually manipulate objects within the scene. To overcome this limitation, we employ mesh representation for objects and adopt the cascade stylization over each object in the scene achieving the consistent stylization as depicted in Figure \ref{fig:pipeline}. Our pipeline starts by sampling objects either user specified or reconstructed from a single-view image provided by the user. Secondly, the text prompt containing style information is used to generate a styled reference scene image as global guidance. The prompt is also used to control the viewpoint-dependent stylized texturization iteratively. What’s more, the previous textured mesh is used to supervise the following mesh texturization. The whole texturization is in a cascaded manner to achieve the multi-object style consistency. Subsequently, the objects are positioned and scaled within the scene based on ChatGPT learnt positions reasoning. Finally,
the final scene is composed.

\subsection{Preliminary}\label{sec:preli}
\paragraph{Text-to-Image Diffusion} Diffusion model \cite{Rombach_2022_CVPR} mainly generates target data sampling from noise (sampled from a simple distribution) by predicting noise. The diffusion model is divided into two processes, diffusion process and reverse process. Both diffusion process and reverse process are parameterized Markov Chains \cite{norris1998markov} or non-Markov Chains \cite{song2022denoising}. The input image $x_{0}$ is first encoded into a latent code $z_{0}$ before the diffusion process. In the forward process, $z_{t}$ is only related to $z_{t-1}$ at the previous moment. This process is regarded as a Markov process and satisfies:
\begin{equation}\label{eq1}
    q\left(z_{1: T} \mid z_{0}\right)=\prod_{t=1}^{T} q\left(z_{t} \mid z_{t-1}\right)
\end{equation}

\begin{equation} \label{eq2}
    q\left(z_{t} \mid z_{t-1}\right)=\mathcal{N}\left(z_{t}, \sqrt{1-\beta_{t}} z_{t-1}, \beta_{t} \mathbf{I}\right)
\end{equation}
Among them, $\beta_{t}$ with different $t$ is predefined and gradually increases from time $1\sim T$.
DDPM \cite{ho2020denoising} uses neural network $ p_{\theta}\left(z_{t-1} \mid z_{t}\right)$ to fit the inverse process $ q\left(z_{t-1} \mid z_{t}\right)$. Finally, 
 $\mu_{\theta}$ is fitted through the neural network:
\begin{equation}
    \mu_{\theta}\left(z_{t}, t\right)=\frac{1}{\sqrt{\alpha}_{t}}\left(z_{t}-\frac{\beta_{t}}{\sqrt{1-\bar{\alpha}_{t}}} \epsilon_{\theta}\left(z_{t}, t\right)\right)
\end{equation}
where $\alpha_{t}=1-\beta_{t}$ , $\hat{\alpha}_{t}=\prod_{t=1}^{T} \alpha_{i}$ , and $\epsilon_{\theta}$ is is a noise predictor, we can learn $\epsilon_{\theta}$ by:
\begin{equation}
    \ell=\mathbb{E}_{t,z_0,\mathbf{\epsilon}}\left[\|\mathbf{\epsilon} - \mathbf{\epsilon}_\theta (\sqrt{\bar{\alpha}_t} z_0+\sqrt{1-\bar{\alpha}_t}\mathbf{\epsilon}, t)\|_2\right]
\end{equation}
where  $\mathbf{\epsilon}$ is a random variable sampled from $\mathcal{N}(\mathbf{0},\mathbf{I})$.

\paragraph{Depth-Aware Control} Currently, the desired result image can be generated through the Depth2Image \cite{rombach2021highresolution} model, given the depth map and text prompt. Through depth control, 3D relationship between light and shadow can be observed in 2D images, which helps to achieve relatively high consistency under multiple viewing angles. However, since the Depth2Image model generates the entire image, when stylizing the mesh, we need to generate stylized textures on the mesh surface at different viewing angles by using an inpainting mask to guide the sampling process. Masks can provide explicit hints about which areas should be generated or kept. By injecting the generation mask $\mathcal{M}$ into the sampling steps, the known regions of the input are denoised. This mask explicitly blends the noised latent code $z_{t}$ and the denoised latent estimate $\hat{z}_{t}$ as follows:
\begin{equation}
    \hat{z}_t = \hat{z}_t \odot \mathcal{M} + z_t \odot (1 - \mathcal{M}).
\end{equation}

\subsection{3D Object Mesh Generation}
\label{sec:mesh_gen}
Creating meshes manually is limited in terms of the number of types and diversity. Thanks to generative models \cite{liu_syncdreamer_2023,liu2023meshdiffusion}, there is a significant enhancement in the variety of objects that can be generated. This is particularly evident in single-view image reconstruction, where obtaining mesh models for real-world objects becomes easier.

The recent work presented by Zero123 \cite{shi_zero123_2023} showcased the ability to generate convincing new perspectives from a single-view image of an object. However, this approach faced challenges in maintaining consistency in both geometry and colors across the generated images. In contrast, SyncDreamer \cite{liu_syncdreamer_2023} has addressed this issue by achieving synchronization through a 3D-aware feature attention mechanism. This mechanism correlates corresponding features across different views, employing a synchronized multi-view diffusion model to capture the joint probability distribution of multi-view images. Consequently, SyncDreamer enables the generation of multi-view-consistent images through a single reverse process.

\begin{figure}[t]
\begin{center}
\includegraphics[width=\columnwidth]{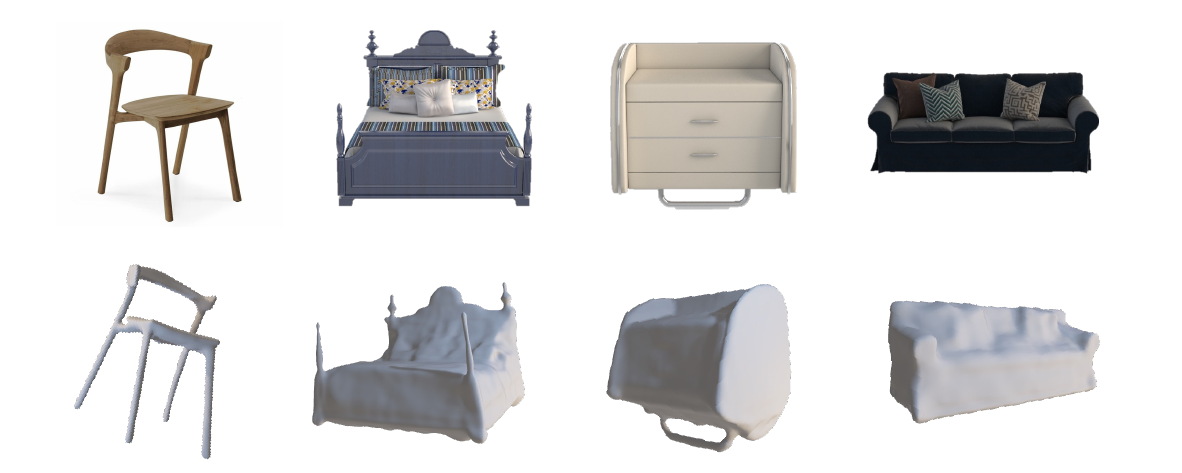}
\end{center}
\caption{The reconstructed meshes from the single-view images of a wooden chair, a bed, a small cabinet and a sofa.}
\label{fig:mesh}
\end{figure}

Given a single-view image and the predefined viewpoints as \(x(1)_0 , ..., x(N)_0\) SyncDreamer learns the joint distribution of all these views \(p\theta(x_0^{(1:N)}|y) := p\theta (x_0^{(1)} , ..., x_0^{(N)}|y)\). It servers as the N synchronized noise predictor by correlating the multi-view features using a 3D-aware attention scheme, which can enforce consistency among multiple generated views. Some generated example meshes are depicted in Figure \ref{fig:mesh} with their corresponding input view accordingly.

\begin{figure*}[t]
\begin{center}
\includegraphics[width=\textwidth]{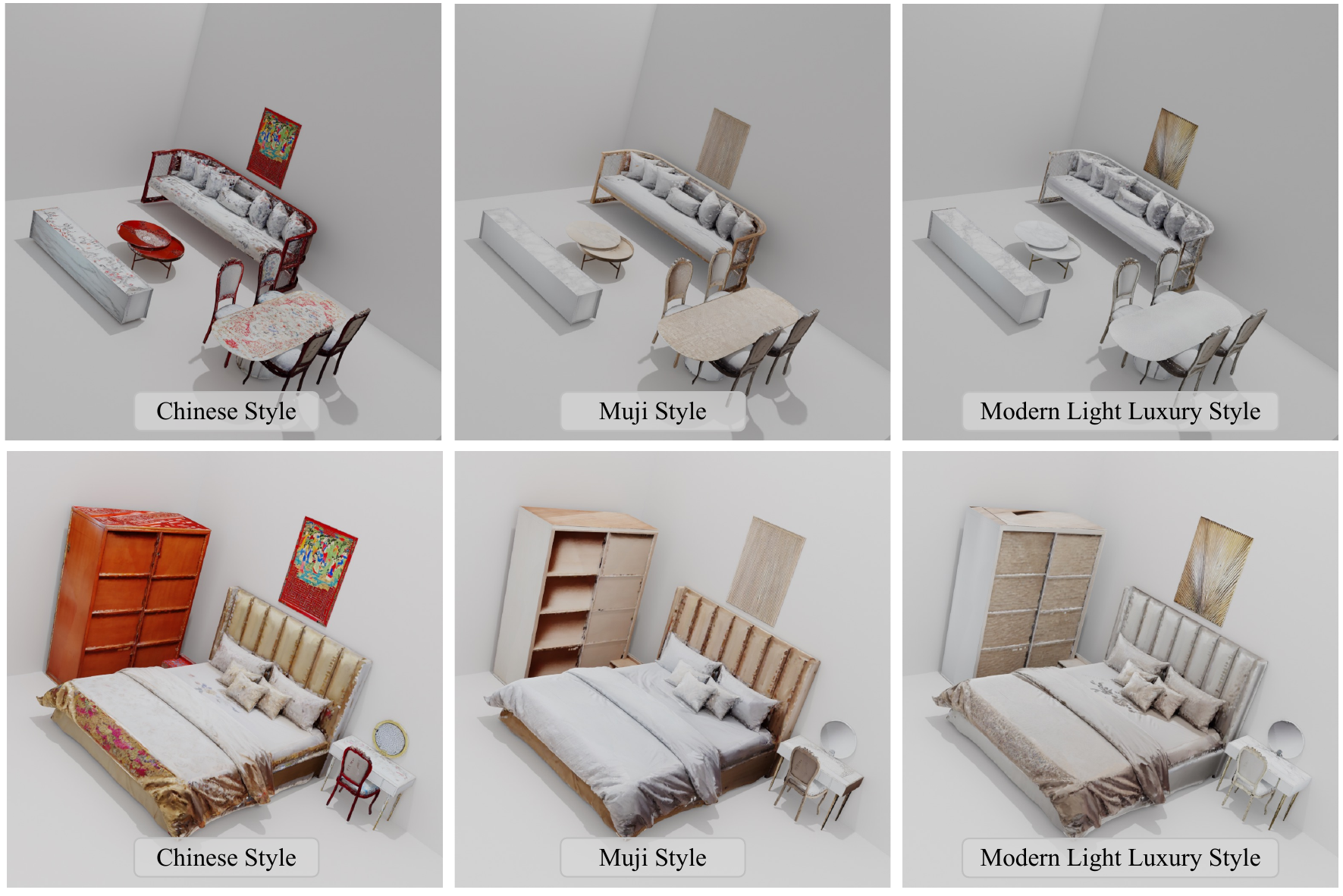}
\end{center}
\vspace{-5mm}
\caption{The diverse stylized 3D indoor scene synthesis using different prompts. The figures in first row depict the typical living room scenes and those in the second row are the bedrooms. The figures in the first column is conditioned by \texttt{Chinese Style}. The objects and camera view are different from what is shown in \ref{fig:teaser}. The figures in the second and third column are conditioned by \texttt{Muji Style} and \texttt{Modern Light Luxury Style}. The placements and geometries of these objects are the exactly same whereas the styles are totally different, meaning our pipeline can fully de-couple geometry and appearance.}
\label{fig:results}
\end{figure*}

\subsection{3D Indoor Scene Stylization}\label{sec:scene_styliz}
We propose to stylize the 3d indoor scene in the auto-regressive way to achieve the multi-object style consistency. To this end, we employ an cascaded way to stylize each object within the indoor scene.

To stylize the meshes obtained from Section \ref{sec:mesh_gen}, we formulate this task as a mesh inpainting task. Recent work Text2tex \cite{chen_text2tex_nodate} employs an iterative process to generate images from various viewpoints, guided by predefined perspectives and supervised by the depth map. The generated image is then utilized to texture the mesh from its corresponding viewpoint. The subsequent viewpoint image is partially painted based on the prior view and the depth map, framing the task as a completion in the subsequent step. In this context, the initial front view image plays a pivotal role as it establishes the overall texture stylization for the target object.

While solely relying on a text prompt offers limits supervision for image generation and also introduces ambiguity in the generated image, potentially compromising style consistency across different objects. To address this, we propose to employ dual modality supervision for scene stylization. Initially, the first mesh inpainting is solely guided by the text prompt. Simultaneously, we output the initial image generated as the complete scene supervision, along with object-level images from various viewpoints. Subsequently, the following mesh is supervised using both the text prompt and the complete scene image, as well as the object-level images. These images are encoded using the CLIP \cite{radford2021learning} image encoder and cross-attentioned with the text feature, providing robust style supervision.

\begin{figure*}[h]
\begin{center}
\includegraphics[width=0.8\linewidth]{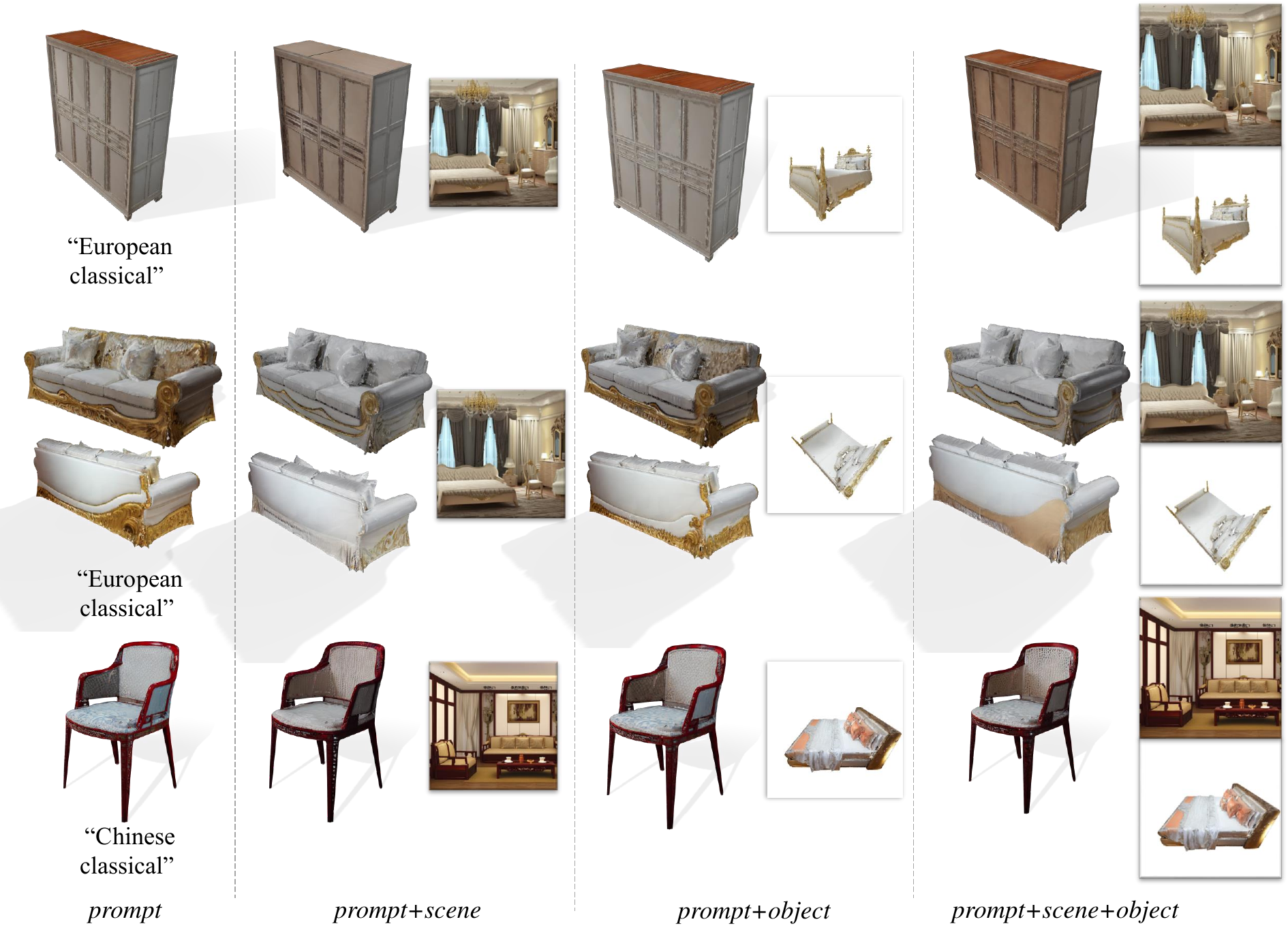}
\end{center}
\vspace{-5mm}
\caption{The stylized mesh guided by prompt (first column), prompt and whole scene images (second column), prompt and object level images (third column) and the whole scene images and object level images (last column).}
\label{fig:ablation}
\end{figure*}

Figure \ref{fig:ablation} illustrates the stylized outcomes achieved through various modality controls. The images in the first column are solely guided by text prompts, those in the second column are guided by both text prompts and the initial whole scene image generated in the first stage, and those in the last column are guided by text prompts, multi-view objects, and the initial image. Upon careful observation, it is evident that the images in the last column exhibit the most consistent style and maintain faithful visual quality.

\subsection{3D Scene Synthesis}\label{sec:scene_synth}
Recently, some methods for synthesizing 3D scenes rely on 360 panoramic pictures \cite{tang_mvdiffusion_2023, yang_dreamspace_nodate} for 3D mesh generation. Even though these approaches often yield visually pleasing results because of the robust diffusion backbone, the generated scenes are confined to the central area of the room, and they inherently struggle with inter-object occlusion and the objects in the scene are not de-coupled, making it impossible for the users to manipulated individual objects.

We align with prevailing 3D scene object placement methods \cite{xu_discoscene_2022, fang_ctrl-room_2023} in our pipeline. These methods utilize 3D bounding boxes as the guide to position the corresponding 3D objects within the scene. This approach allows for more comprehensive control of the scene, addressing the limitations of those methods solely relied on panoramic pictures. What's more, based on the exisitng bounding boxes in 3D-FRONT \cite{fu20203dfront}, we instruct and guide ChatGPT \cite{OpenAI_GPT3} to learn the inter position of indoor objects through in-context learning and generate the relevant placement bounding boxes. We mainly focus on regular shape living-rooms and bedrooms. In this way, users are free to generate different sences according their specific requirement. The prompt turning are placed in the supplementary material.

Guided by the bounding boxes, each object is automatically positioned at its corresponding location and scaled properly. If the mesh object is not in its default canonical position, the viewpoint of the object would be adjusted accordingly. Several synthesized scenes are presented in Section \ref{sec:exp}, accompanied by a detailed analysis.



\section{Experiments}
\label{sec:exp}
\paragraph{Dataset and Baselines.} The experiments are conducted using the 3D-FRONT dataset \cite{fu20203dfront}, an indoor scene dataset that includes 6.8K houses and 140K rooms. Specifically, our focus is on living rooms and bedrooms with commonly acceptable furniture placement for the given task. We also compare our method with scene stylization approaches such as \cite{tang_mvdiffusion_2023}, \cite{yang_dreamspace_nodate}, \cite{yu_texture_nodate}, and \cite{hollein_stylemesh_2022}.

\begin{figure}[t]
\begin{center}
\includegraphics[width=0.9\columnwidth]{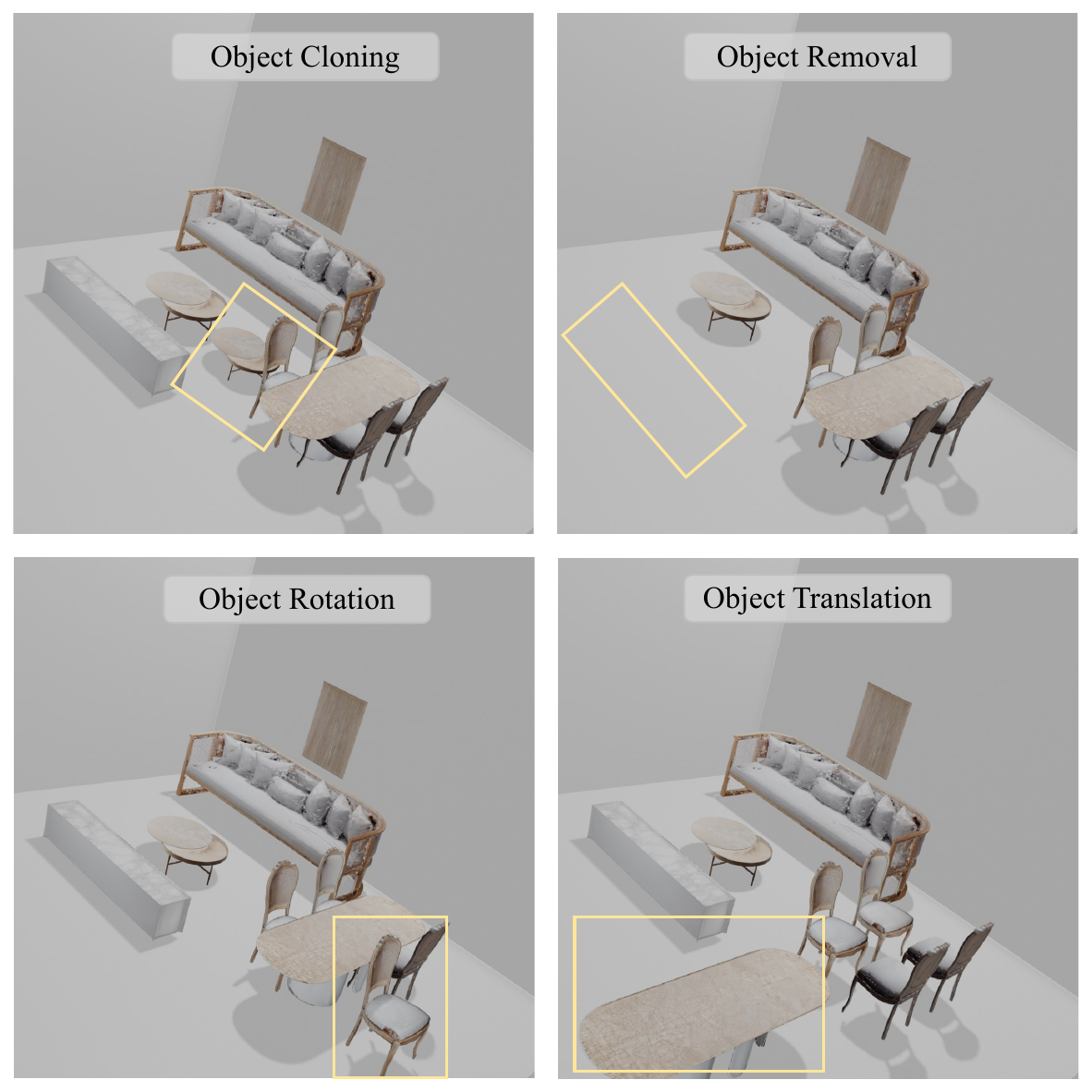}
\end{center}
\vspace{-5mm}
\caption{We perform different user controls on scene objects, such as rearrangement, removal and cloning. The origin image is the living room with \texttt{Muji Style} as in \ref{fig:results}.}
\label{fig:ctrl}
\end{figure}

\subsection{Controllable Scene Generation}
The bounding boxes incorporated into our model provide versatile user controls over scene objects. In the following sections, we assess the flexibility and effectiveness of our model by applying various 3D manipulation techniques. Examples of these manipulations are illustrated in Figure \ref{fig:ctrl}. User can control the rotation and translation of the objects in the scenes without affecting their appearance by controlling the corresponding bounding boxes. Transforming shapes in Figure \ref{fig:ctrl} shows consistent results. In particular, with one chair rotated, the rest shapes do not change, suggesting desired multi-view consistency. Our model can also properly handle mutual occlusion. Users can update the scene such as removing or cloning existing object by copying and pasting a box to a new location. Explicit camera control is also permitted. Rendered image from rotating the camera randomly are depicted in Appendix.


\subsection{Comparison on Generative 3D Scenes}
\paragraph{Experiment setting.} We evaluate our method by comparing it with both 3D scene synthesis and scene-level mesh stylization works both quantitatively and qualitatively.

\begin{figure}[t]
\begin{center}
\includegraphics[width=\columnwidth]{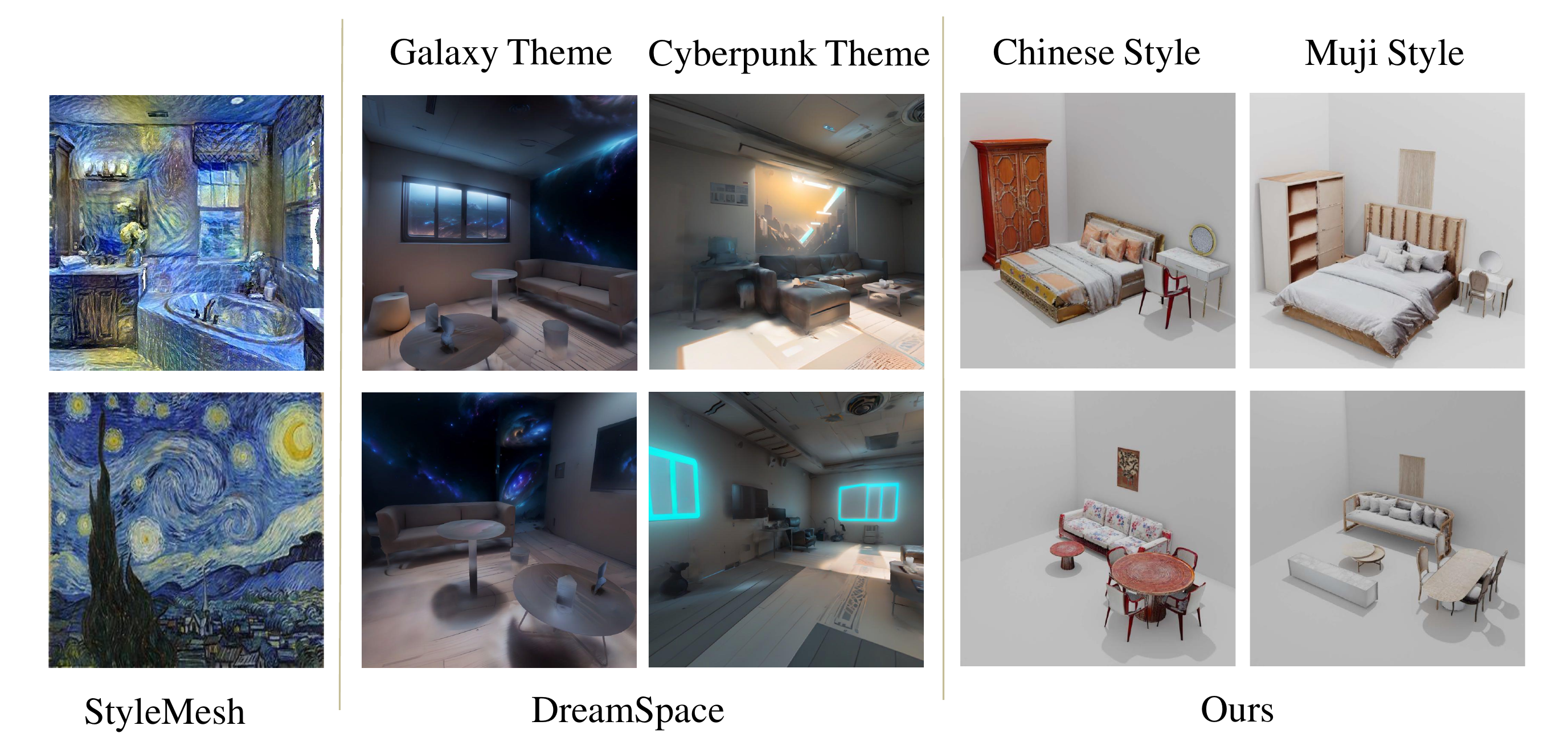}
\end{center}
\vspace{-5mm}
\caption{Visual comparison of text/image-guided stylized texture generation. We present the scene-level mesh stylization results for StyleMesh and DreamSpace under style image and text prompts, respectively (rendered view through a fixed camera perspective).}
\label{fig:qualitative comparison}
\end{figure}

\paragraph{Qualitative Comparison.}  We visualize the qualitative comparison stylized scene results in Figure \ref{fig:qualitative comparison} and \ref{fig:results} where we both exhibit the overview mesh rendering views and corresponding text prompt. We synthesis a variety of 3D indoor scenes with distinct styles using different prompts. The \texttt{Chinese Style} scenes are rendered  using different camera poses from those depicted in Figure \ref{fig:teaser}. The scenes in the second and third columns are synthesized using prompt to the \texttt{Muji Style} and \texttt{Modern Light Luxury Style} respectively. Despite maintaining identical object placements and geometries, the styles are entirely distinct, demonstrating the capability of our pipeline to effectively separate the geometry and appearance aspects. In terms of global and local style consistency, our approach achieves uniformity of style across the entire scene. In the case of StyleMesh \cite{hollein_stylemesh_2022}, where only the style image is used as control information and there is no high-level semantic prior, the global style is influenced solely by the appearance, resulting in a stylized output that is typically chaotic and lacks meaningful texture. For DreamSpace \cite{yang_dreamspace_nodate}, while semantic information is retained through panoramic scene texturing, the generated panoramic textures using the 2D diffusion model result in distortions in alignment  and texture propagation. Consequently, the resulting stylized scenes suffer from texture blurring, artifacts and may lack clear distinguish ability. Clearly, our methodology amalgamates more coherent textures with pristine and more abundant local intricacies. This leads to superior outcomes in terms of both global and local style consistency by referencing the preceding stylized object and incorporating a global style image as an supplementary control, all while upholding semantic information. Additionally, distinctive style attributes and differentiation are apparent in stylized scenes generated by different text prompts.

\begin{table}[t!]
\centering
\resizebox{1.0\linewidth}{!}{
\tabcolsep 2pt
\begin{tabular}{lcccccc}
\toprule
\multicolumn{1}{c}{\multirow{2}{*}{Methods}} & \multicolumn{2}{c}{Quantitative Metrics} & \multicolumn{2}{c}{User Study} \\ \cmidrule(lr){2-3} \cmidrule(lr){4-5}  
\multicolumn{1}{c}{} & \multicolumn{1}{l}{CLIP Score $\uparrow$} & \multicolumn{1}{l}{Aesthetic $\uparrow$} & \multicolumn{1}{l}{Correctness $\uparrow$} & \multicolumn{1}{l}{Quality $\uparrow$} \\ \midrule
StyleMesh~\cite{hollein_stylemesh_2022} & 0.184 & 4.812 & 2.68 & 2.76 \\
MVDiffusion~\cite{tang_mvdiffusion_2023} & 0.174 & 4.263 & 1.37 & 1.49 \\
TEXTure~\cite{yu_texture_nodate} & 0.187 & 5.265 & 2.57 & 2.20 \\
DreamSpace \cite{yang_dreamspace_nodate} & 0.214 & \textbf{5.771} & 3.38 & 3.55 \\  \midrule
Ours & \textbf{0.245} & \textbf{5.671} & \textbf{3.55} & \textbf{3.88} \\
\bottomrule
\end{tabular}}
\vspace{-3mm}
\caption{
We perform quantitative evaluation and user studies on output 3D indoor scene for StyleMesh~\cite{hollein_stylemesh_2022}, MVDiffusion~\cite{tang_mvdiffusion_2023}, TEXTure~\cite{yu_texture_nodate} DreamSpace \cite{yang_dreamspace_nodate} and our method.  
}
\label{tab:compare}
\end{table}

\paragraph{Quantitative Comparison.} For the quantitative assessment, we employ a similar evaluation method as in DreamSpace \cite{yang_dreamspace_nodate}. We adopt the CLIP Score \cite{radford2021learning} to see how well the generated views match the given text prompts. Additionally, we use an aesthetic scoring method introduced by LAION \cite{schuhmannclip+}. As shown in Table \ref{tab:compare}, our approach gets the highest scores CLIP Score and comparable aesthetic score to DreamSpace. This shows that our created texture closely fits the given text prompts and maintains high quality.

\subsection{User Study} 
We carried out a user study to evaluate our method in comparison to others. Twenty users were given the task of organizing rendered views from textured meshes produced by various methods, focusing on two aspects: the correctness of image-text matching and perceptual quality. Participants assigned scores based on their rankings, with a score of 4 given to the top-ranked method and a score of 1 for the lowest-ranked one. Our method receives the highest preferences by a substantial margin, highlighting the remarkable visual quality and the degree of image-text matching achieved by our approach.

\subsection{Ablation Study}

\paragraph{Text prompt guidance.} Owing to the inherent semantic ambiguity within textual information, we observed that relying solely on text guidance for generating stylized textures led to inconsistencies across multiple objects. The identical text could introduce ambiguities, yielding distinct stylization outcomes for different objects and thereby causing incongruent styles among them.
 
\paragraph{Cascaded object direct stylization.} We note that the direct stylization approach, cascading across the object, enables the mitigation of style inconsistencies for each object, thereby enhancing the quality of appearance. This improvement is facilitated by the newly generated object referring the overall style of the preceding object. Each object contributes valuable style perceptions, leading to a consistent stylistic coherence across the entire scene.

\paragraph{Global condition image guidance.} Despite limited guidance and supervision solely through text prompts and other object styles, achieving a high degree of style consistency across the entire scene remains challenging. To address this, we introduce global condition image guidance to oversee and regulate both the global scene and its constituent objects. Through this global condition image guidance, the entire scene and its objects attain uniformity in style, resulting in superior visual quality.

\section{Conclusion}
In summary, we present an novel 3D indoor scene synthesis pipeline that is tailored to produce distinct mesh objects using either text prompts or single-view images. The objects within these scenes can undergo stylization using either text instructions or a designated style image, thereby maintaining a cohesive style throughout various objects. Our pipeline illustrates that the resultant indoor scenes display visual harmony in both style and spatial organization, presenting a unified and visually appealing composition.

\subsection{Limitations and Future works}
Despite the advancements, there are still some limitations of our pipeline. Firstly, the style supervision from the whole scene still need to be further explored. Secondly, It will provide advantages to incorporate some optimization algorithm on the objects arrangement, like LEGO-Net \cite{wei_lego-net_nodate}. Furthermore, the aesthetic quality of the synthesis of 3D indoor scenes remains an under-explored area. It will be beneficial with aesthetic score to enhance the overall scene synthesis visual quality.    
{
    \small
    \bibliographystyle{ieeenat_fullname}
    \bibliography{main}
}


\end{document}